\documentclass{bmvc2k}

\usepackage{subfiles}
\usepackage{CJKutf8}
\usepackage{bm}
\usepackage{amsmath}
\usepackage{booktabs}
\usepackage{caption}
\usepackage{multirow}
\usepackage{float}
\usepackage{amsfonts}
\usepackage{caption}
\usepackage{subcaption}

\title{Reading Chinese in Natural Scenes with a Bag-of-Radicals Prior}

\addauthor{Yongbin Liu}{liuyongbin@buaa.edu.cn}{1}
\addauthor{Qingjie Liu}{qingjie.liu@buaa.edu.cn}{1, 2}
\addauthor{Jiaxin Chen}{jiaxinchen@buaa.edu.cn}{1}
\addauthor{Yunhong Wang}{yhwang@buaa.edu.cn}{1, 2}

\addinstitution{
 School of Computer Science and Engineering\\
 Beihang University\\
 Beijing, China
}
\addinstitution{
 State Key Laboratory of Virtual Reality Technology and System\\
 Beihang University\\
 Beijing, China
}

\runninghead{Liu et al.}{Reading Chinese in Natural Scenes}

\begin{document}

\maketitle

\begin{abstract}
Scene text recognition (STR) on Latin datasets has been extensively studied in recent years, and state-of-the-art (SOTA) models often reach high accuracy. However, the performance on non-Latin transcripts, such as Chinese, is not satisfactory. In this paper, we collect six open-source Chinese STR datasets and evaluate a series of classic methods performing well on Latin datasets, finding a significant performance drop. To improve the performance on Chinese datasets, we propose a novel radical-embedding (RE) representation to utilize the ideographic descriptions of Chinese characters. The ideographic descriptions of Chinese characters are firstly converted to bags of radicals and then fused with learnable character embeddings by a character-vector-fusion-module (CVFM). In addition, we utilize a bag of radicals as supervision signals for multi-task training to improve the ideographic structure perception of our model. Experiments show performance of the model with RE + CVFM + multi-task training is superior compared with the baseline on six Chinese STR datasets.
\end{abstract}

\section{Introduction}
\noindent The scene text recognition (STR) task is to recognize text in natural scenes, such as road signs, shop banners, house numbers, etc, making it attractive for many downstream applications, for instance, mobile language translators, autonomous driving and geo-location. Unlike traditional optical character recognition (OCR) that mainly aims at document recognition, the shapes of text in natural scene images are arbitrary, such as curved shapes; and more challenging are that they may be captured under low illumination, various distances, and fonts are diverse.

In recent years, extensive efforts have been devoted to scene text recognition \cite{shi2016end,wang2017gated,borisyuk2018rosetta,li2019show,yue2020robustscanner}. These methods mainly focus on Latin, which are trained and tested on Latin datasets. Few works have considered recognizing east Asian texts such as Chinese, Japanese, and Korean, which have distinct structures to Latin characters, and nearly 20\% of world population are using them.

\begin{figure}[tb]
    \centering
    \includegraphics[width=.8\textwidth]{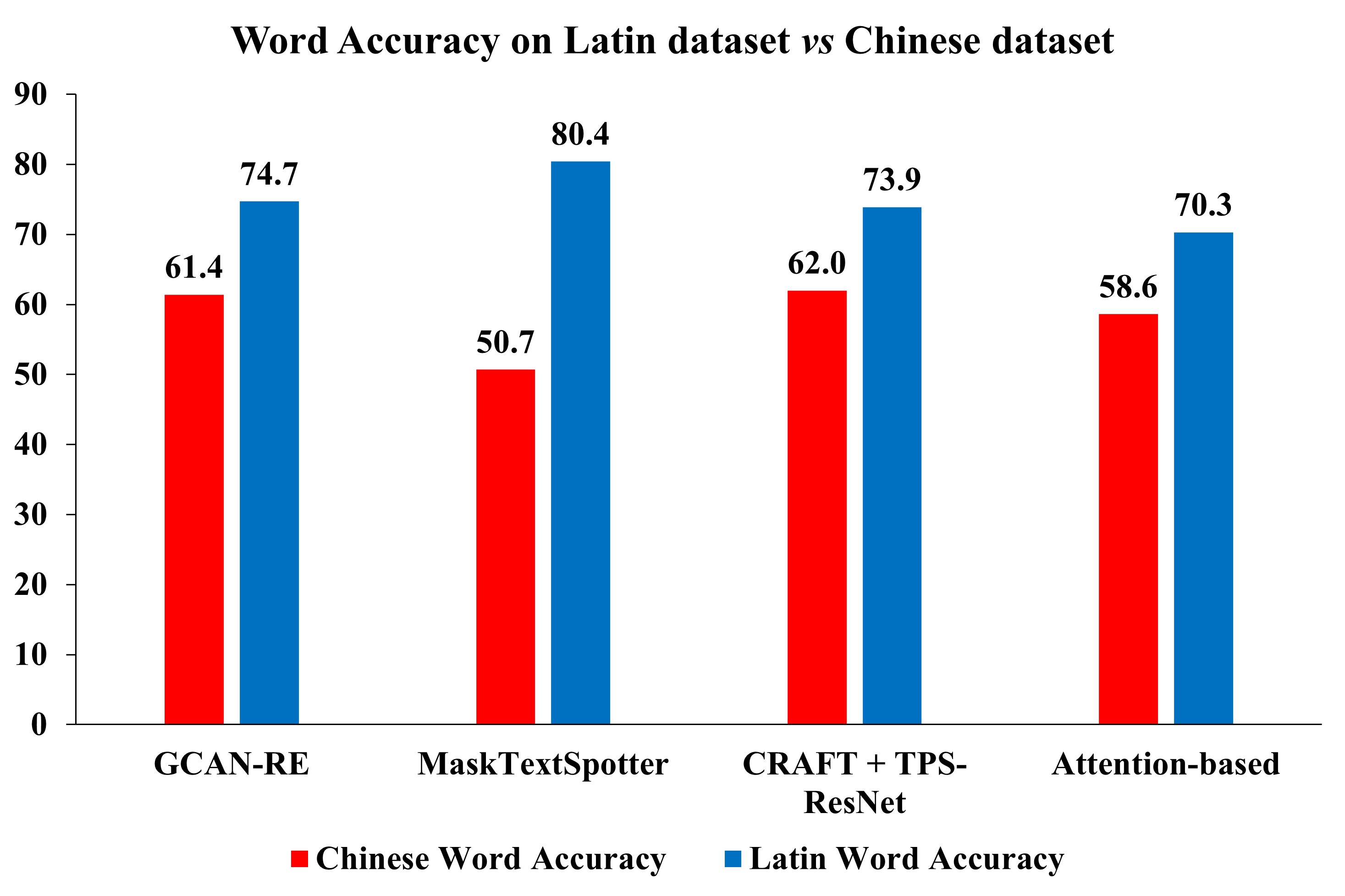}
    \caption{The performance of top-4 methods on Latin part and Chinese part of ICDAR-2019 Robust Reading Challenge on Arbitrary-Shaped Text leaderboard (scene text recognition task). There is a significant performance gap between Latin and Chinese recognition in natural scenes.}
    \label{figure:cn_vs_latin}
\end{figure}

Fig. \ref{figure:cn_vs_latin} shows the word accuracy on Latin transcripts and Chinese transcripts on ICDAR-2019 Robust Reading Challenge on Arbitrary-Shaped Text leaderboard \cite{chng2019icdar2019}, and it is obvious that there is a significant performance gap between Chinese part and Latin part. The word accuracy on Latin transcripts reach 70+\%, while on Chinese transcripts is only approximately 60\%, which is far lower.

To facilitate research in text recognition in natural scene images, ICDAR organized several competitions on multi-language scene text reading, including LSVT \cite{sun2019icdar}, ArT \cite{chng2019icdar2019} and MLT \cite{nayef2019icdar2019}. These tracks release datasets that contain Chinese transcripts. However, Chinese samples are mixed together with other languages, making it difficult to assess performance on Chinese transcripts. Furthermore, the methods in competitions use different configurations of experiments and comparisons between them are unfair. In order to make a fair comparison between different approaches and benchmark Chinese character recognition in natural scenes, we collect several widely used Chinese STR datasets and evaluate several approaches on them with the same configuration. Since Latin transcripts are also common used in Chinese Street views, such as `SPA' (balneotherapy), `KTV' (Karaoke TV) and numbers on road signs, we evaluate the models on both Latin part and Chinese part, separately.

A model that performs well on Latin transcripts may have low performance on Chinese transcripts, indicating that we need to develop new models to read Chinese characters. Inspired by the learning process of pupils on Chinese characters, we utilize the radical information of Chinese characters to improve the performance of Chinese STR.

\begin{figure}[tb]
        \centering
        \includegraphics[width=.6\textwidth]{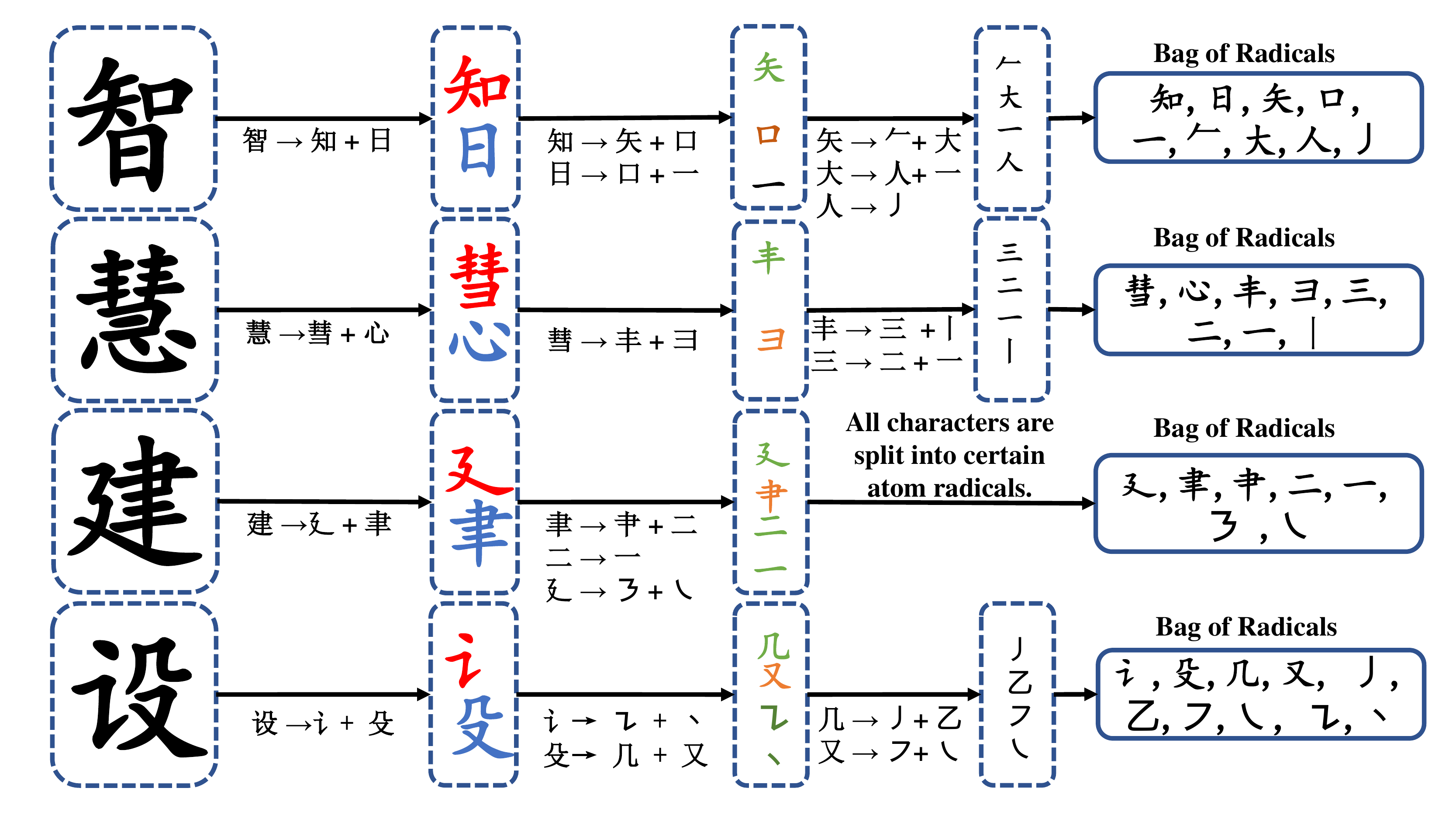}
        \caption{This figure demonstrates how we transform a Chinese character to a bag of radicals. Firstly, we split the character into several basic Chinese radicals, then we collect all radicals of this character to a bag of radicals.}
        \label{fig:bag_of_radicals}
\end{figure}

The main contributions of this work are three-fold:
\begin{enumerate}
    \item We evaluate 10 well-known STR models on six large-scale Chinese STR datasets. This result provides a fair comparison in Chinese STR field.
    \item We find a performance gap between the Chinese recognition and the Latin recognition. And even worse, the models performing very well on Latin datasets may have a low performance on Chinese datasets. It indicates that simply deploying a Latin STR method in Chinese scenes may cause a performance drop.
    \item A novel Chinese STR method that fuses the bags of radicals with CVFM is proposed, and we design a multi-task radical classification branch to improve the character structural perception. Experiments demonstrate that our method is superior to the baseline model. 
\end{enumerate}

\section{Related Work}
\subsection{Scene Text Recognition}
Early STR methods recognize words in a bottom-up manner \cite{yao2014strokelets}. With the development of deep learning techniques, researchers resort to convolutional neural networks (CNNs) as feature extractors and update them in a data driven manner. The optimized visual features are then reformulated by recurrent neural networks (RNN) as sequential features. The CNN and RNN are commonly defined as the encoder, transforming the original images into hidden representations. There are two families of decoders that convert the features into probabilistic logits, that is CTC decoder \cite{shi2016end,liu2016star,borisyuk2018rosetta} and 1D attention-based decoder \cite{lee2016recursive,shi2016robust,baek2019wrong}. Both CTC and 1D attention decoders frame the classification process as seq2seq learning like natural language processing (NLP) tasks.

 Observing that the 1D attention decoder misses pixel-wise information and attention, \cite{li2019show} proposes a novel 2D attention decoder (SAR), which significantly improves the recognition performance on Latin datasets. Gaussian constrained attention networks (GCAN) \cite{qiao2021gaussian} utilizes a learnable gaussian constrained module to improve the qualities of attention masks. 

\subsection{Chinese Character Recognition with Radicals as Priors}
Several works have noticed the use of Chinese character radicals as a kind of knowledge prior. \cite{zhang2018radical} and \cite{wang2018denseran} propose DenseRAN, a single-character recognition network that firstly predicts the structure code of a Chinese character image and then transforms the code to a Chinese character. The decoding mechanism of \cite{zhang2018radical} and \cite{wang2018denseran} is designed for \textit{single-character} recognition, and for a transcript that contains several Chinese characters, the radical code would be extremely long, resulting in impracticable word accuracy. \cite{wang2019radical} uses Chinese radicals as a kind of structural prototypes for few shot character recognition. Similar to DenseRAN, \cite{wang2019radical} also decomposes Chinese characters into basic radicals and encodes a character image into a radical sequence. \cite{wang2017radical} proposes to recognize Chinese characters by multi-label classification. \cite{wang2017radical} mainly focuses on printed single characters, and cannot be directly used to recognize scene text words. \cite{chenzero} proposes a stroke-based priors for Chinese character recognition. It first uses visual recognizer to encode the characters into stroke sequences, and then lookup from a lexicon for the most similar character as the prediction. Actually it is not a radical-based method since a radical consists several basic strokes.


Most above methods are designed for \textit{single-character} recognition and only conduct experiments on \textit{single-character} datasets. Their application in scene text recognition is intractable due to the long radical coding. In our method, we take bags-of-radicals for two usages, the first one is to use it as a embedding feature to fuse with visual features; the other is using them as supervision signals to build up a structure-awareness branch. The usage of radicals of our method is novel and different from \cite{ma2008new}, \cite{cha2020few}, \cite{wang2001optical}, \cite{wu2019joint},  \cite{wang2017radical}, \cite{wang2019radical}, \cite{park2020few}, \cite{wang2018denseran}, \cite{chenzero} and \cite{zhang2018radical}.

\section{The Proposed Approach}

\begin{figure}{h}
    \centering
    \includegraphics[width=\textwidth]{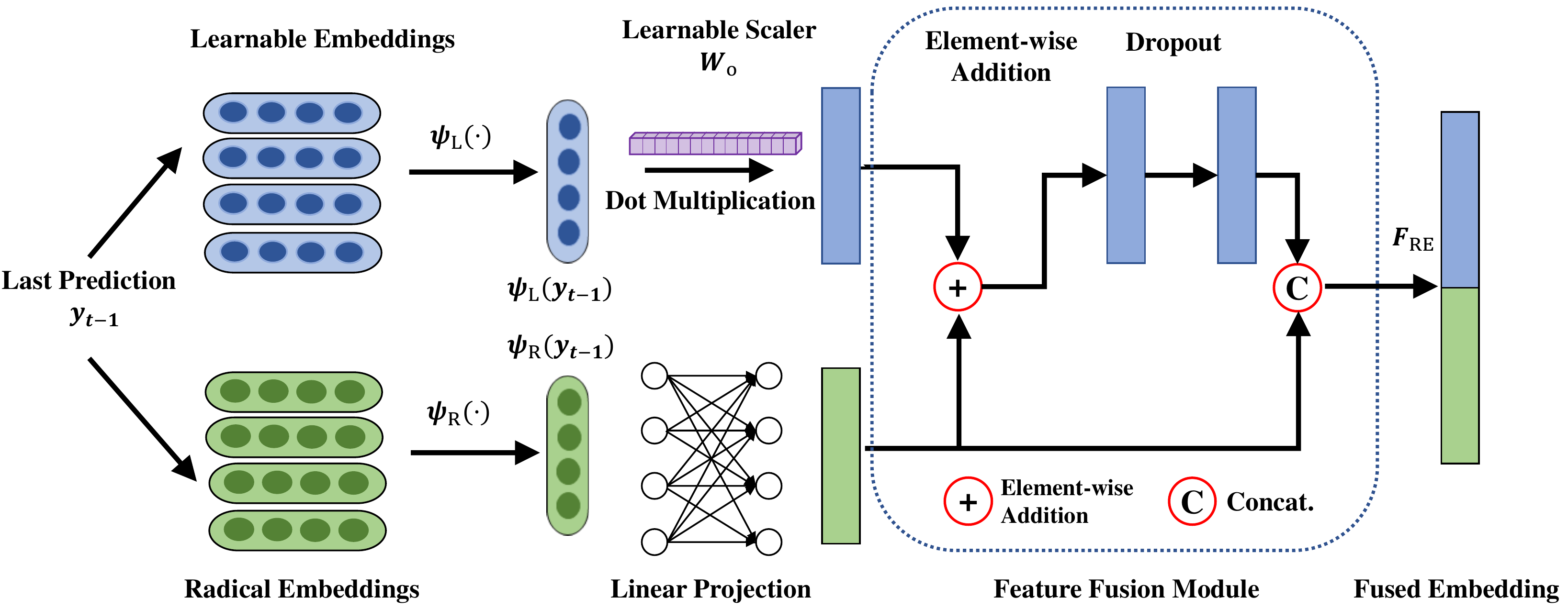}

    \caption{The pipeline of our character vector fusion module. Firstly the looked up learnable embedding of $y_{t-1}$ is scaled by a learnable weight $\bm{W}_o$ to $\bm{W}_o \cdot \bm{o}$, and the radical embedding is linearly transformed by $\bm{W}_r$ to $\bm{W}_r \bm{r}$, which is projected into the same vector space as $\bm{W}_o \cdot \bm{o}$. These two features are summed and sent to a dropout layer, which aims to protect the features from overfitting. The result is concatenated with radical embedding $\bm{r}$, and that is the fused embedding feature $\bm{F}_{RE}$.}
    \label{figure:CVFM}
\end{figure}

In this section, we describe details about the framework and the key components of the proposed method. 
\subsection{Framework Overview}

\begin{figure}[tb]
    \centering
        \includegraphics[width=.9\textwidth]{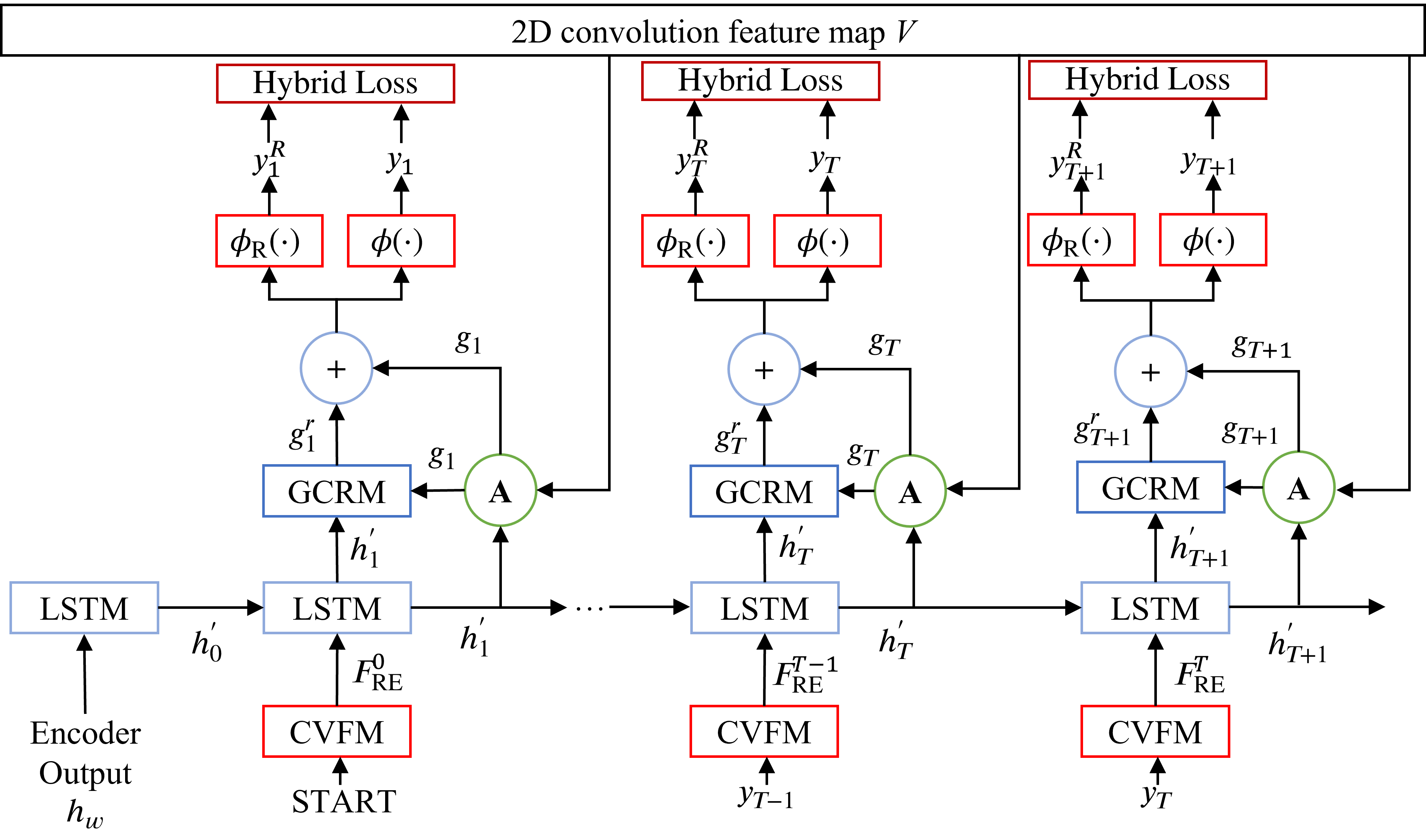}
        \caption{The structure of the decoder used in our model. The contextual feature $h_t^{'}$, visual feature map extracted by backbone networks $V$ and the embedding vector of last predicted character $\psi(y_{t-1})$ are fused by 2D attention modules. The GCRM module is to use Gaussian masks \protect\cite{qiao2021gaussian} to constrain the attention map. \textcircled{A} is 2D attention,  and \textcircled{+} means element-wise addition.}
    \label{figure:GCAN_graph}
\end{figure}

Our method follows the same pipeline as GCAN \cite{qiao2021gaussian}, while the main difference lies in the decoder part. As illustrated in Fig.~\ref{figure:GCAN_graph}, given an input image, the 2D visual feature map $\bm{V}$ is  extracted by a ResNet backbone. $\bm{h}_w$ is the contextual information encoded by an LSTM cell. The decoder predicts the characters in a sequential step-by-step manner, where $\bm{h}'_i$ indicates the contextual information. Specifically, at time step $T$, the precedent prediction $\bm{y}_{T-1}$ is firstly fed into the proposed Character Vector Fusion Module (CVFM) and transformed into a character representation $\bm{F}_{RE}^{T-1}$ with the Bag-of-Radicals prior, which is subsequently taken as the input of an LSTM cell to generate the contextual knowledge $\bm{h}_T^{\prime}$. $\bm{h}_T^{\prime}$ is thereafter integrated with the visual map $\bm{V}$ by passing a  2D attention module \textcircled{A} with an output $\bm{g}_{T}$. $\bm{g}_{T}$ is further fused with $\bm{h}_T^{\prime}$ by a Gaussian Constrained Refinement Module (GRCM), yielding a refined vector $\bm{g}_{T}^{r}$, which is ensembeled with $g_{T}$ by element-wise addition. The final representation $\bm{F}_{T}$ is used to predict the character $\bm{y}_{T}$ via a predictor $\phi(\cdot)$. To exploit the radical prior for learning a discriminative $\bm{F}_{T}$, we employ an auxiliary task, \emph{i.e.} predicting which radicals exist in the current character, by adopting a radical classification head $\phi_{R}(\cdot)$. Accordingly, a radical-induced hybrid loss is developed to optimize the overall network.


\subsection{Character Vector Fusion Module}
\label{subsection:radical_embedding}
In this section, we will describe the CVFM module. Before that, we firstly employ the Bag-of-Radicals Embedding for an arbitray Chinese character. 

\textbf{Bag-of-Radicals Embedding.} Given a Chinese character, we utilize the open source Unicode dictionary \textbf{\emph{cjkvi}} \cite{kawabata2013} to progressively decompose it into a bag of radicals. For instance, as shown in Fig. \ref{fig:bag_of_radicals}, the character \begin{CJK*}{UTF8}{gbsn} `慧' \end{CJK*} can be firstly divided into two principle parts, \emph{i.e.,} \begin{CJK*}{UTF8}{gbsn} `彗' \end{CJK*} and \begin{CJK*}{UTF8}{gbsn} `心' \end{CJK*}. The radical \begin{CJK*}{UTF8}{gbsn} `彗' \end{CJK*} can be further separated into \begin{CJK*}{UTF8}{gbsn} `彐' and `丰'\end{CJK*}. The decomposition process as above continues until the \begin{CJK*}{UTF8}{gbsn} `丰'\end{CJK*} is decomposed into \emph{atom} radicals \begin{CJK*}{UTF8}{gbsn} `三', `二', `一'\end{CJK*} and `$~| $~'. By this mean, an arbitrary Chinese character can be represented by a bag of radicals as well as its atom radicals. For instance, the bag-of-radicals representation of \begin{CJK*}{UTF8}{gbsn}`慧'\end{CJK*} is \begin{CJK*}{UTF8}{gbsn} \{彗,心,丰,彐,三,二,一,$~~| $~~\}\end{CJK*}

By performing the bag-of-radicals (BoR) decomposition on all characters in the Chinese charset (totally 5,941 characters), we can obtain a bank of 1,637 distinct basic Chinese radicals $CRBank=\{CR_1, CR_2, CR_3, \cdots, CR_D \}$, where $D$=1637. Based on $CRBank$, given an arbitrary Chinese character $C$, we encode it into a 1,637 dimensional BoR vector $\bm{r}_{C}=[v_1,v_2,\cdots, v_{1637}]\in \mathbb{R}^{1637}$ according to its BoR decomposition $\{CR_{i_1},CR_{i_2},\cdots, CR_{i_K}\}$, where $v_{j}=1$ for $j=i_1,i_2,\cdots,i_K$, and $0$ otherwise. As for any Latin character $C_{L}$ that does not contain Chinese radicals, we set the BoR of it a zero vector. By encoding all 5,941 characters into the BoR vectors, we can obtain an matrix $\bm{R} = [\bm{r}_{1};~\cdots;~ \bm{r}_{5941}]\in \mathbb{R}^{5941 \times 1637}$.

\textbf{CVFM.} We now introduce the character vector fusion module (CVFM) which integrates the geographic information. It first converts the predicted character at the last time step $\bm{y}_{T-1}$ to radical embeddings (RE) and then fuses RE with a learnable embedding (LE) to obtain fusion $\bm{F}_{RE}$. The decoder uses the fused features to compute contextual information by an LSTM cell, as shown in Fig. \ref{figure:GCAN_graph}, and further makes prediction. 
The fusion steps are described as follows:
\begin{align}
        \bm{o} &= \psi_{L}(\bm{y}_{t-1}) \\
        \bm{r} &= \psi_{R}(\bm{y}_{t-1}) \\
        \bm{F}_{RE} &= \text{Concat}(\text{Dropout}(\bm{W}_r \bm{r} + \bm{W}_o \cdot \bm{o});\bm{r})
\end{align}

 Firstly, the looked up learnable embedding of $\bm{y}_{t-1}$ is scaled by weight $\bm{W}_o$ to obtain $\bm{W}_o \cdot \bm{o}$, and the radical embedding is linearly transformed by $\bm{W}_r$ to the same dimensionality as $\bm{o}$. A dropout layer is applied to avoid overfitting. The result is concatenated with radical embedding $\bm{r}$, finally obtaining the fused embedding feature $\bm{F}_{RE}$. This mechanism is shown in Fig. \ref{figure:CVFM}. We use the linear projection to transform the radical embedding into the same semantic space as learnable embedding, and the dot multiplication is only to adjust the scale of learnable embedding.
 
 In the baseline (GCAN), the decoder only uses a learnable embedding to extract contextual knowledge, which does not contain any structural prior. Our modification is to utilize the radical prior to boost the learning of contextual features.
 
 \subsection{A Radical-Induced Hybrid Loss}
In addition to the use of radical embeddings, we also build a multi-task branch to boost the character-structure awareness of our model. In the prediction head of the decoder, we use the bags-of-radicals as ground truth to supervise the model. The hybrid loss in our model is defined as follows:
\begin{align}
    \hat{\bm{y}}_{T} &= \phi({\bm{g}_{T}}) \\
    \hat{\bm{y}}_{T}^{R} &= \phi_{R}(\bm{g}_{T}) \\
    \mathcal{L}_{O} &= {\rm CrossEntropy}(\hat{\bm{y}}_{T}, \bm{y}_{T}) \\
    \mathcal{L}_{R} &= {\rm BCE}(\hat{\bm{y}}_{T}^{R}, \bm{y}_{T}^{R})\\
    \mathcal{L} &= \mathcal{L}_{O} + \mathcal{L}_{R}
\end{align}
where $\bm{g}_T$ represents the glimpse vector in Fig. \ref{figure:GCAN_graph}. $\hat{\bm{y}}_{T}$ represents the probability logits of the prediction head of the baseline, and the corresponding ground truth $ \bm{y}_{T} $ is a one-hot vector. $\phi_{R}$ is the radical prediction classifier, and the output is $ \hat{\bm{y}}_{T}^{R}$. The ground truth of the radical branch $\bm{y}_{T}^{R}$ is the proposed radical embeddings. We use cross entropy loss (CE loss) and binary cross-entropy loss (BCE loss) to optimize the two branches respectively, and the losses are summed up to the hybrid loss. It is worth noting that the radical representations of Latin characters are null and not taken into BCE loss computation. Specifically, we use a mask to label zero for the Latin characters and one for Chinese characters, and the mask are multiplied by the loss matrix, and we demonstrate how this mask works in supplementary materials.

\section{Experiments}
\label{section:exp}
In this section, we conduct experiments to show the superiority of using radical embeddings, CVFM and the proposed hybrid loss for reading Chinese in natural scenes. We compare our model with baselines and conduct performance analysis on the evaluation results. We also utilize ablation study to verify each component of our model. The \textit{case study}, \textit{performance analysis on Latin subset} and \textit{compatibility with text detectors} are demonstrated in supplementary materials.

\subsection{Dataset Details}

\setlength\tabcolsep{3.0pt}

\begin{table}[!htbp]
\begin{center}
    \caption{The number of train and test samples in each Chinese dataset.}
    \label{table:dataset_statistics}
    \scalebox{0.76}{\begin{tabular}{cccccc}
    \toprule
    \multirow{2}[4]{*}{Dataset} & \multirow{2}[4]{*}{Year} & \multicolumn{2}{c}{Train} & \multicolumn{2}{c}{Test} \\
\cmidrule{3-6}          &       & Total & Non-Latin & Total & Non-Latin \\
    \midrule
    ArT   & \multicolumn{1}{c|}{2019} & 39,493 & 11,280 & 9,748 & 2,798 \\
    CASIA & \multicolumn{1}{c|}{2018} & 57,599 & 41,091 & 25,097 & 17,889 \\
    ctw   & \multicolumn{1}{c|}{2018} & 176,472 & 176,448 & 12,562 & 12,560 \\
    LSVT  & \multicolumn{1}{c|}{2019} & 213,985 & 176,814 & 24,143 & 20,097 \\
    RCTW  & \multicolumn{1}{c|}{2017} & 34,727 & 26,901 & 9,017 & 6,990 \\
    ReCTS & \multicolumn{1}{c|}{2019} & 86,755 & 67,697 & 22,033 & 17,079 \\
    \midrule
    Total &       & 609,031 & 500,231 & 102,600 & 77,413 \\
    \bottomrule
    \end{tabular}}%

\end{center}
\end{table}

\setlength\tabcolsep{3pt}
\begin{table*}[ht]
\centering
\caption{We evaluate models including CRNN \protect\cite{shi2016end}, RARE \protect\cite{shi2016robust}, R2AM \protect\cite{lee2016recursive}, STAR-NET \protect\cite{liu2016star}, GRCNN \protect\cite{wang2017gated}, Rosetta \protect\cite{borisyuk2018rosetta}, TRBA \protect\cite{baek2019wrong}, SAR \protect\cite{li2019show}, RobScan \protect\cite{yue2020robustscanner}, GCAN \protect\cite{qiao2021gaussian} on all six datasets using word accuracy and 1-N.E.D metrics.}
\label{table:benchmark_result}
\begin{subtable}{\textwidth}
    \centering
    \scalebox{0.76}{
    \begin{tabular}{c|ccccccc|ccccccc}
    \toprule
    \multirow{2}[4]{*}{Model} & \multicolumn{7}{c|}{Word accuracy on all transcripts} & \multicolumn{7}{c}{1-Normalized edit distance on all transcripts} \\
\cmidrule{2-15}          & ArT   & CASIA & ctw   & LSVT  & RCTW  & ReCTS & All   & ArT   & CASIA & ctw   & LSVT  & RCTW  & ReCTS & All \\
    \midrule
    GRCNN & 42.1  & 40.5  & 31.1  & 37.5  & 41.5  & 53.9  & 41.7  & 72.5  & 69.5  & 61.3  & 64.8  & 67.5  & 72.6  & 68.2  \\
    R2AM  & 60.1  & 53.6  & 53.2  & 58.7  & 55.5  & 67.6  & 58.5  & 78.9  & 74.9  & 69.8  & 75.5  & 73.9  & 80.6  & 75.9  \\
    CRNN  & 44.4  & 44.9  & 38.5  & 45.7  & 45.8  & 60.9  & 47.8  & 75.0  & 72.9  & 65.3  & 70.6  & 70.9  & 78.3  & 72.6  \\
    Rosetta & 64.6  & 59.0  & 54.8  & 62.2  & 61.3  & 74.6  & 63.3  & 84.5  & 81.6  & 76.8  & 81.3  & 81.1  & 86.9  & 82.3  \\
    RARE  & 72.1  & 64.6  & 62.8  & 68.0  & 66.6  & 76.9  & 68.7  & 86.5  & 82.6  & 78.2  & 82.7  & 81.9  & 86.7  & 83.3  \\
    STAR-NET & 59.5  & 52.9  & 49.5  & 54.6  & 56.1  & 71.0  & 57.7  & 80.6  & 76.4  & 69.0  & 74.1  & 75.9  & 84.1  & 77.0  \\
    TRBA  & 75.6  & 66.0  & 63.8  & 69.3  & 68.1  & 78.7  & 70.3  & 88.9  & 83.3  & 78.7  & 83.4  & 82.9  & 88.0  & 84.3  \\
    RobScan & 78.1 & 68.8  & 66.8  & 72.1  & 70.3  & 80.6  & 72.9  & \textbf{89.9} & 85.8  & 82.4  & 86.0  & 85.0  & 89.4  & 86.5  \\
    SAR   & 74.8  & 69.0  & 65.2  & 71.3  & 69.9  & 80.8  & 72.2  & 88.2  & 85.5  & 81.6  & 85.3  & 84.5  & 89.5  & 86.0  \\
    GCAN  & 75.4  & 70.0  & 65.2  & 72.1  & 70.8  & 81.4  & 72.8  & 88.6  & 86.2  & 81.5  & 85.8  & 84.9  & 89.8  & 86.4  \\
    \midrule
    Ours  & \textbf{78.2}  & \textbf{71.3} & \textbf{67.5} & \textbf{73.9} & \textbf{72.5} & \textbf{82.3} & \textbf{74.6} & 89.8  & \textbf{86.5} & \textbf{82.5} & \textbf{86.5} & \textbf{85.6} & \textbf{90.2} & \textbf{87.0} \\
    \bottomrule
    \end{tabular}}%

\end{subtable}

\end{table*}

\textit{ICDAR-2019 ArT(ArT)} is proposed in \cite{chng2019icdar2019} to evaluate the STR models on texts with complex shapes, such as curved shapes. There are 49,241 cropped images in the training dataset. Since the test dataset does not provide labels, we randomly split the training set into two parts. 20\% (9,748 items) of them is used as testing, and 80\% (39,493 items) serves as the training set. Among all characters, 25.8\% are Chinese and the rest of 74.2\% are Latin.

\textit{CASIA-10k (CASIA)} is proposed by \cite{he2018multi} for evaluating scene text detections. Bounding boxes are annotated with text content and cropped into samples. Finally, there are 57,599 samples in the training set and 25,097 in the test set. 

\textit{Large Chinese Text Dataset in the Wild (ctw)} is proposed by \cite{yuan2019ctw}, mainly containing street view images captured by cameras. Based on bounding boxes, there are 176,472 cropped images for training, and 12,562 for test. 

\textit{Large-scale Street View Text (LSVT)} is a large-scale street view text dataset \cite{sun2019icdar}. There are 30k images which contain bounding boxes and transcripts. 213,985 cropped images are used in the training part and 24,143 are used for the test part. 

\textit{ICDAR2017 Competition on Reading Chinese Text in the Wild (RCTW)} \cite{shi2017icdar2017} is created for the ICDAR-2017 competition. Since the ground truth files of test datasets are not publicly available, we firstly crop the images in the training dataset and obtain 43,744 samples. 20\% (9,017) are randomly selected to compose our test dataset and the remaining 34,727 samples for training. 

\textit{Reading Chinese Text on Signboard (ReCTS)} is created by \cite{zhang2019icdar} for ICDAR 2019 STR. Each image is annotated with bounding boxes of rectangles. Since the test dataset is not publicly available, we randomly divide the training dataset into two parts. 86,755 samples (80\%) are used as training samples and the other 22,033 images are for evaluation. 

We also demonstrate the number of non-Latin samples in each dataset in Tab. \ref{table:dataset_statistics} (\textit{non-latin} samples contain at least one Chinese character).

\vspace{-10pt}
\subsection{Experimental Setups}
We build up the STR model pipeline using MMOCR \cite{mmocr2021} and deep STR benchmark \cite{baek2019wrong}. To make fair comparisons, all models adopt the same experimental setting and are trained on the same training data, that is ArT + CASIA + ctw + LSVT  + RCTW + ReCTS (609,031 samples in total, as shown in Tab. \ref{table:dataset_statistics}). The evaluation datasets are the test part of above six datasets (102,600 samples in total). During training, we adopt the Adam optimizer \cite{kingma2014adam}, with an initial learning rate 0.001 and no weight decay. The batch size during training is 1024, distributed on eight GTX 2080Ti GPUs. The height of each input image is rescaled to 128 pixels while keeping the height/weight ratio invariant; if the new width is greater than 256 pixels, we resize the image to  128 $\times $ 256, otherwise pad  it to 128 $\times$ 256 with zeros. Data augmentation is applied during training, including color jitter transformation (brightness 0.5, contrast 0.5, saturation 0.5, hue 0.5), the random affine transformation (degrees 0.5, translate (0.1, 0.1), scale (0.95, 1)), the random rotation transformation (degrees 5), and the identity transformation. For each image sample, we randomly select one of the above augmentation types in each mini-batch.

\subsection{Evaluation Protocols}
The common metric to evaluate the similarity of two transcripts is word accuracy (two strings are considered the same if and only if every two aligned characters in the string pair are the same). In order to eliminate the influence of symbols and spaces, previous STR works ignore all symbols (!@\#\$\%, etc), and we follow the practice in this work.

Another measure in ICDAR competitions \cite{chng2019icdar2019} is to use 1 - Normalized Edit Distance(1-N.E.D) to evaluate the similarity of two strings. The computation  is $1-\frac{\text{EditDis}(s_1, s_2)}{\text{max}(\text{len}(s_1), \text{len}(s_2))}$. In our experimental results, we report model performance using both measures.

\subsection{Experimental Results}
We select 10 well known STR models to conduct experiments, including CRNN \cite{shi2016end}, RARE \cite{shi2016robust}, R2AM \cite{lee2016recursive}, STAR-NET \cite{liu2016star}, GRCNN \cite{wang2017gated}, Rosetta \cite{borisyuk2018rosetta}, TRBA \cite{baek2019wrong}, SAR \cite{li2019show}, RobScan \cite{yue2020robustscanner} and GCAN \cite{qiao2021gaussian}. These models can be divided into three categories according to decoders they use: CTC-based (CRNN, GRCNN, Rosetta, STAR-NET), 1D attention-based (R2AM, RARE, TRBA), and 2D attention-based (SAR, RobScan, GCAN). The experimental results are shown in Tab. \ref{table:benchmark_result}. Based on the experimental results, we reach the following conclusions:
\begin{itemize}
    \item The models that adopt 2D attention decoders outperform models with other types of decoders. The lowest average word accuracy of 2D attention models on non-latin transcripts is 71.3, achieved by SAR. It is higher than any other methods using 1D attention decoders or CTC decoders such as TRBA (68.5) and Rosetta (63.7).
    \item Our method outperforms other models significantly. As evaluated on all transcripts, our method outperforms the baseline GCAN by 1.8\% word accuracy and the largest margins are ArT (2.8\%) and ctw (2.3\%), indicating that our method works on both curved text and street view text images.
\end{itemize}

\subsection{Ablation Study}

In order to verify the function of our modules, we conduct an ablation study. We first remove the radical binary cross-entropy loss (BCE) and then the CVFM module. As shown in Tab. \ref{table:GCAN_RE_performance}, using radical embeddings and fusion modules can improve the performance by 1.5\% on non-Latin transcripts. The additional BCE loss can improve the performance by another 0.3\%. The experimental results validate the effectiveness of CVFM and BCE loss.

We also conduct ablation study w.r.t each component of the CVFM module. We remove the scaler, dropout operation and set radical embedding to zero respectively. The experimental results are shown in Tab. \ref{table:ablation_CVFM}. According to the results, removing any component or set the radical embedding causes significant performance drop, hence we validate the necessity of each component in CVFM.

\setlength\tabcolsep{2.5pt}
\begin{table*}[hpbt]
\centering
\caption{Ablation study w.r.t CVFM and BCE loss of our method. The first line shows the performance of our baseline (GCAN). The radical embeddings (RE) and CVFM are then added and the performance is shown in the 2nd line. The 3rd line demonstrates the effectiveness of our BCE Loss.}

    \scalebox{0.7}{\begin{tabular}{ll|cccccccc|cccccccc}
    \toprule
    \multicolumn{2}{c|}{Ablation} & \multicolumn{8}{c|}{Word accuracy on all transcripts}         & \multicolumn{8}{c}{Word accuracy on non-Latin transcripts} \\
    \multicolumn{1}{p{3.3em}}{CVFM} & \multicolumn{1}{p{3.3em}|}{BCE} & ArT   & CASIA & ctw   & LSVT  & RCTW  & ReCTS & All   &  Gain & ArT   & CASIA & ctw   & LSVT  & RCTW  & ReCTS & All   & Gain \\
    \midrule
          &       & 75.4  & 70.0  & 65.2  & 72.1  & 70.8  & 81.4  & 72.8  & \textcolor[rgb]{ .439,  .678,  .278}{/} & 74.7  & 68.1  & 65.2  & 71.6  & 70.7  & 80.5  & 71.8  & \textcolor[rgb]{ .439,  .678,  .278}{/} \\
    \checkmark &       & 77.4  & 71.0  & 67.6  & 73.6  & 72.2  & 82.1  & 74.3  & \textcolor[rgb]{ .439,  .678,  .278}{+1.5 } & 78.1  & 69.0  & 67.6  & 73.3  & 72.0  & 81.2  & 73.2  & \textcolor[rgb]{ .439,  .678,  .278}{+1.4 } \\
    \checkmark & \checkmark & 78.2  & 71.3  & 67.5  & 73.9  & 72.5  & 82.3  & 74.6  & \textcolor[rgb]{ .439,  .678,  .278}{+1.8 } & 79.2  & 69.3  & 67.5  & 73.6  & 72.5  & 81.5  & 73.5  & \textcolor[rgb]{ .439,  .678,  .278}{+1.7 } \\
    \bottomrule
    \end{tabular}}%

\label{table:GCAN_RE_performance}
\end{table*}

\begin{table*}[hpbt]
    \caption{Ablation w.r.t components of the CVFM.}
    \label{table:ablation_CVFM}

      \scalebox{0.76}{\begin{tabular}{c|ccccccc|ccccccc}
    \toprule
    \multirow{2}[2]{*}{Ablation for CVFM} & \multicolumn{7}{c|}{Word Accuracy on All Transcripts} & \multicolumn{7}{c}{Word Accuracy on non-Latin Transcripts} \\
          & ArT   & CASIA & ctw   & LSVT  & RCTW  & ReCTS & All   & ArT   & CASIA & ctw   & LSVT  & RCTW  & ReCTS & All \\
    \midrule
    full CVFM  & 77.4  & 71.0  & 67.6  & 73.6  & 72.2  & 82.1  & 74.3  & 78.1  & 69.0  & 67.6  & 73.3  & 72.0  & 81.2  & 73.2  \\
    w/o scaler & 70.4  & 66.1  & 65.2  & 69.6  & 68.1  & 79.7  & 70.3  & 71.3  & 64.7  & 65.2  & 69.3  & 68.3  & 79.0  & 69.7  \\
    w/o dropout & 76.6  & 69.4  & 66.4  & 72.3  & 70.9  & 81.1  & 73.0  & 77.2  & 67.0  & 66.4  & 71.9  & 70.7  & 80.1  & 71.8  \\
    RE set zero & 72.9  & 69.3  & 67.6  & 72.4  & 71.0  & 82.1  & 73.0  & 72.8  & 67.7  & 67.6  & 72.1  & 71.1  & 81.5  & 72.4  \\

    \bottomrule
    \end{tabular}}%
\end{table*}

\section{Conclusion}
In this work we collect six Chinese scene text datasets to evaluate the previous models performing well on Latin datasets fairly. From the experimental results we can conclude that the 2D-attention models reach high word accuracy on both Latin datasets and Chinese datasets.
In order to further improve the performance of Chinese STR, we propose RE, CVFM and hybrid radical losses for multi-task training. The experiments on the Chinese STR benchmark show that our method is superior compared with our baseline (GCAN) and other methods on six datasets.

\bibliography{egbib}
\end{document}


\maketitle

\section{Non-Latin Transcript Recognition Performance}
In the main paper, we give the recognition performance on all transcripts. In this part, we separately give the prediction accuracy on non-Latin transcripts in Tab. \ref{tab:non_latin}.

\begin{table}[htpb]
    \caption{The table lists the performance of 10 selected models evaluated by word accuracy and 1-normalized edit distance on six datasets (only the non-Latin samples are taken into computation).}
    \label{tab:non_latin}
    \centering
    \scalebox{0.65}{\begin{tabular}{c|ccccccc|ccccccc}
    \toprule
    \multirow{2}[4]{*}{Model} & \multicolumn{7}{c|}{Word accuracy on non-Latin transcripts} & \multicolumn{7}{c}{1-Normalized edit distance on non-Latin transcripts} \\
\cmidrule{2-15}          & ArT   & CASIA & ctw   & LSVT  & RCTW  & ReCTS & All   & ArT   & CASIA & ctw   & LSVT  & RCTW  & ReCTS & All \\
    \midrule
    GRCNN & 42.9  & 42.9  & 31.1  & 37.7  & 44.6  & 51.9  & 41.8  & 72.2  & 70.0  & 61.3  & 64.2  & 68.9  & 70.5  & 67.2  \\
    R2AM  & 63.9  & 51.0  & 53.2  & 57.9  & 55.7  & 66.2  & 57.4  & 78.8  & 72.3  & 69.8  & 74.3  & 72.9  & 78.9  & 74.2  \\
    CRNN  & 52.4  & 46.7  & 38.5  & 46.8  & 47.9  & 60.1  & 48.7  & 78.9  & 72.2  & 65.3  & 70.2  & 70.8  & 76.9  & 71.7  \\
    Rosetta & 65.1  & 60.6  & 54.8  & 62.6  & 63.5  & 74.5  & 63.7  & 86.0  & 81.8  & 76.8  & 81.3  & 81.6  & 86.5  & 82.0  \\
    RARE  & 74.7  & 61.9  & 62.8  & 67.2  & 66.5  & 75.1  & 67.2  & 87.5  & 80.9  & 78.2  & 82.0  & 81.4  & 85.4  & 82.0  \\
    STAR-NET & 51.5  & 53.1  & 49.5  & 53.7  & 57.4  & 69.4  & 56.6  & 75.8  & 75.1  & 69.0  & 72.8  & 75.6  & 82.8  & 75.3  \\
    TRBA  & 77.7  & 62.6  & 63.8  & 68.5  & 67.2  & 77.1  & 68.5  & 89.2  & 81.0  & 78.7  & 82.5  & 81.9  & 86.7  & 82.7  \\
    RobScan & \textbf{85.2} & 67.0  & 66.8  & 71.8  & 70.4  & 79.7  & 72.0  & \textbf{93.6} & 84.6  & 82.4  & 85.7  & 84.9  & 88.6  & 85.8  \\
    SAR   & 74.2  & 66.7  & 65.2  & 70.9  & 69.7  & 79.8  & 71.3  & 87.5  & 84.2  & 81.5  & 84.9  & 84.2  & 88.7  & 85.3  \\
    GCAN  & 74.7  & 68.1  & 65.2  & 71.6  & 70.7  & 80.5  & 71.8  & 88.1  & 85.0  & 81.5  & 85.3  & 84.7  & 89.1  & 85.5  \\
    \midrule
    Ours  & 79.2  & \textbf{69.3} & \textbf{67.5} & \textbf{73.6} & \textbf{72.5} & \textbf{81.5} & \textbf{73.5} & 89.9  & \textbf{85.3} & \textbf{82.5} & \textbf{86.1} & \textbf{85.4} & \textbf{89.5} & \textbf{86.1} \\
    \bottomrule
    \end{tabular}}
\end{table}

\section{Case Study}
\begin{figure}
    \centering
    \begin{subfigure}{.9\textwidth}
    \centering
    \includegraphics[width=\textwidth]{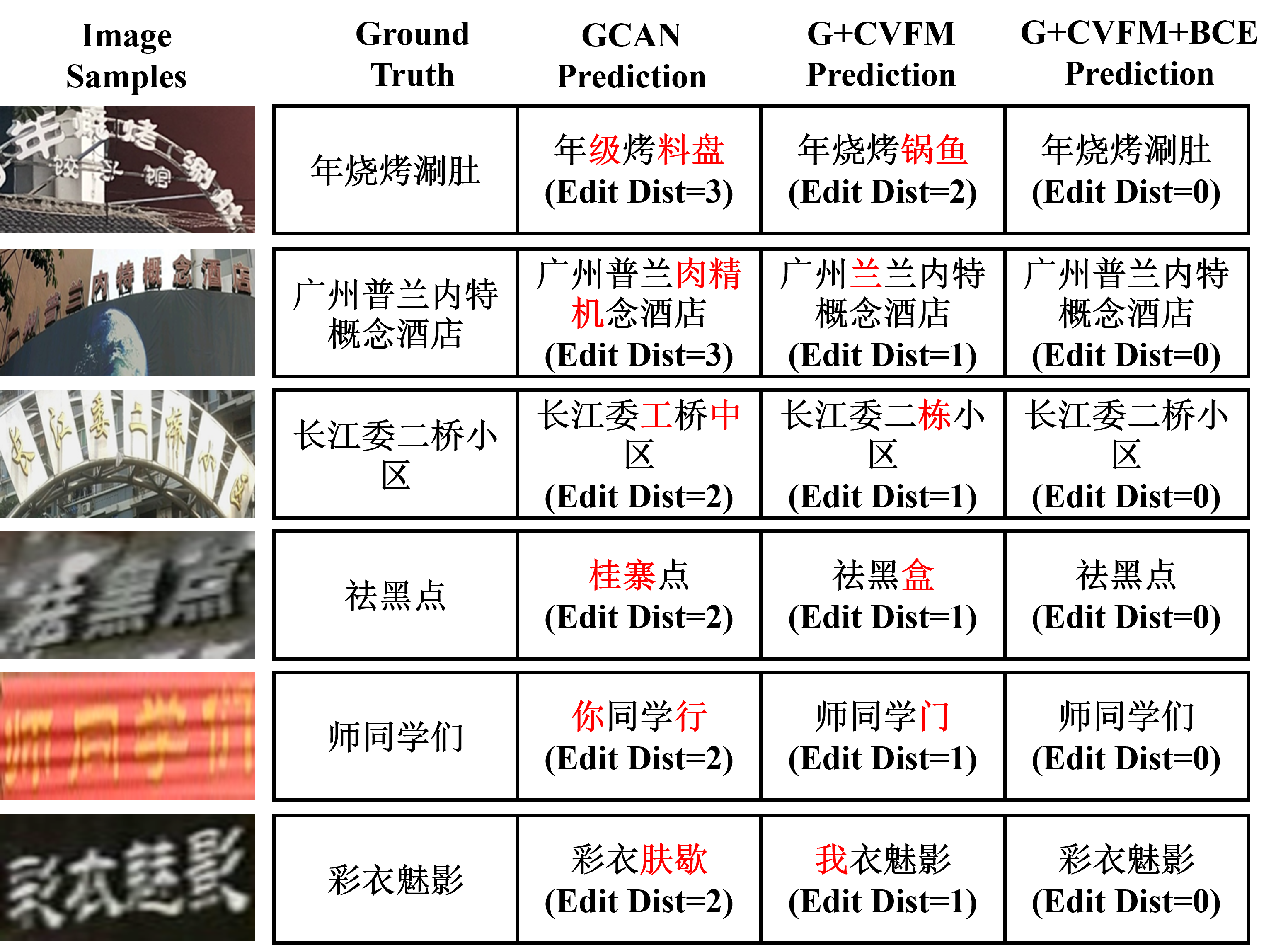}
    \caption{Samples predicted successfully predicted by our model.}
    \label{fig:success}
    \end{subfigure}
    
    \begin{subfigure}{.9\textwidth}
        \centering
    \includegraphics[width=\textwidth]{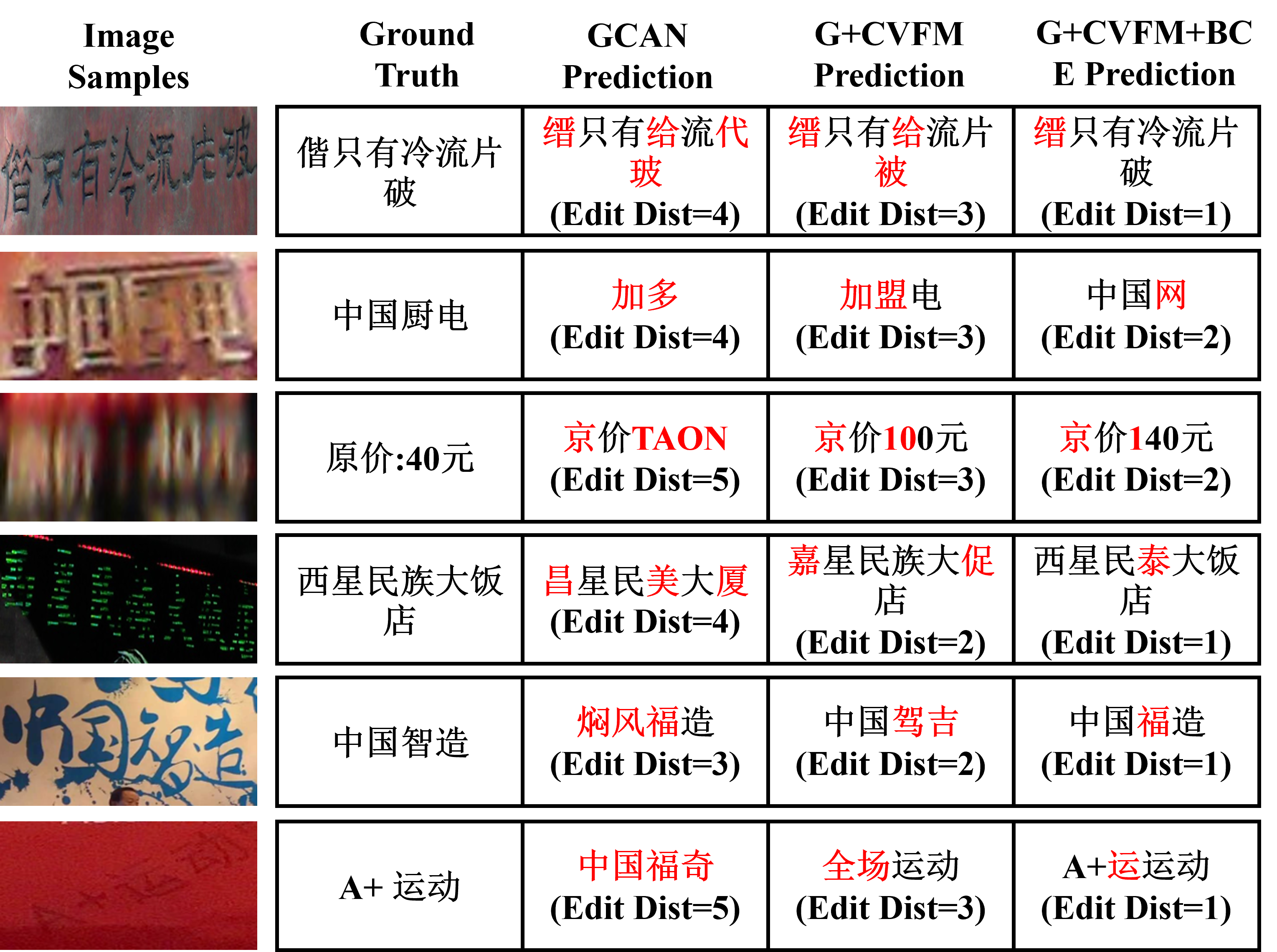}
    \caption{Failure cases. As shown in the figure, the fail reasons include low resolutions, blur, different type and decorated characters.}
    \label{fig:failure}
    \end{subfigure}
    \caption{We give successful cases and failure cases in Fig.\ref{fig:success} and Fig.\ref{fig:failure}, respectively. The wrongly predicted characters are marked as red color.}
 \vspace{-20pt}

\end{figure}

In order to intuitively analyze the improvement of our CVFM module and BCE loss, we give some successful cases predicted by our model and some failure cases to analyze the defect of our model. As shown in Fig. \ref{fig:success}, each line demonstrates a sample, followed by the predictions of GCAN (our baseline model), predictions of GCAN + CVFM module (G+CVFM), and predictions of GCAN + CVFM module + BCE loss (G+CVFM+BCE). The predictions and edit distances between predictions and groud truths are also given. It is quite clear that the CVFM module and BCE loss can improve the structural awareness of the characters. For example, \begin{CJK*}{UTF8}{gbsn}桥\end{CJK*} and \begin{CJK*}{UTF8}{gbsn}栋\end{CJK*} are similar characters, and the model with CVFM + BCE successfully predict it, but GCAN does not (this result is located in line \#3).

Besides successful cases, we also dive into the failure cases and give some reasons that the model fails to recognize. The first line shows that the model fails to recognize a Chinese character of an ancient shape, since the shape of the first character \begin{CJK*}{UTF8}{gbsn}偕\end{CJK*} is in its ancient style, which is different from the modern usage. The third character in the 2nd text line seems very blur, which is the reason causing the failure. The 3rd and 4th line are two examples with low resolution. These samples are difficult to recognize even by human. The 5th example is a decorated Chinese character, and there are some ornamental symbols around them, making difficult to recognize. The last failure case is due to the low contrast ratio, since the foreground color and background color are all red.

\section{Performance Analysis on the Latin Subset}
We now give the model performance on Latin datasets in this section.

The model performance on Latin transcripts achieves 75\%+, and it seems that the results do not outperform the counterpart on non-Latin transcripts (71\%+) by a large margin. We argue it is because the train datasets are not language-wise balanced (only 17\% samples are Latin, as previously described). To validate this, we train GCAN and our model using two experimental settings. The first one is to train models from scratch (GCAN (Latin) and Ours (Latin)) using the pure Latin part of our datasets, and the second is to finetune models given in Tab. 2 (in the main paper) on Latin transcripts (GCAN (FT) and Ours (FT)). The experimental results are listed in Tab. \ref{table:Latin_subset}. The \textit{ctw} column is removed because there are only two Latin samples in the test dataset. Compared with the non-Latin results, the performance on Latin transcripts are much higher. Take GCAN (FT) as an example, the performance on CASIA-Latin is 76.4, 8.3 higher than CASIA non-Latin; and on LSVT-Latin is 78.4, 6.8 higher than LSVT non-Latin. It is worth noting that in these two experiments, the radical embedding is set to zero and no BCE loss is used because there are no radicals for Latin characters.

Another interesting observation is that the performance on Latin transcripts also rises after we add CVFM module. As previously described, the radical embedding of Latin characters is set to zero vector, so only the original learnable embedding is used during training. We argue that the improvement mainly credits to the scaler module operated on the extracted learnable embedding. According to the results in \ref{table:Latin_subset}, adding CVFM brings an performance improvement of approximately 1\%, (GCAN (Latin) \textit{vs} ours (Latin), and GCAN (FT) \textit{vs} ours (FT)). The improvement on the previous Latin benchmark (refer to Tab. \ref{table:Latin_benchmark}) also echoes the argument.

\begin{table}[htpb]
  \caption{Recognition accuracy on pure Latin transcripts.}
  \label{table:Latin_subset}%
  \centering
    \begin{tabular}{c|cccccc}
    \toprule
    \multirow{2}[4]{*}{Model} & \multicolumn{6}{c}{Word Accuracy on Latin Transcripts} \\
\cmidrule{2-7}          & ArT   & CASIA & LSVT  & RCTW  & ReCTS & All \\
    \midrule
    GRCNN & 41.8  & 34.3  & 36.5  & 30.9  & 60.8  & 41.7  \\
    R2AM  & 58.5  & 60.0  & 62.9  & 54.8  & 72.6  & 62.1  \\
    CRNN  & 41.1  & 40.5  & 40.2  & 38.5  & 63.8  & 45.0  \\
    Rosetta & 64.4  & 54.9  & 60.1  & 53.7  & 75.2  & 62.2  \\
    RARE  & 71.0  & 71.4  & 72.2  & 67.2  & 83.3  & 73.4  \\
    STAR-NET & 62.7  & 52.2  & 59.3  & 51.8  & 76.4  & 61.0  \\
    TRBA  & 74.8  & 74.6  & 73.8  & 71.1  & 84.3  & 76.1  \\
    RobScan & 75.3  & 73.3  & 74.0  & 69.8  & 83.8  & 75.7  \\
    SAR   & 75.0  & 74.5  & 73.5  & 70.7  & 84.2  & 76.1  \\
    GCAN  & 75.6  & 74.8  & 74.4  & 71.0  & 84.4  & 76.5  \\
    Ours  & 77.8  & 76.4  & 75.6  & 72.6  & 85.1  & 78.1  \\
    \midrule
    GCAN (Latin) &   76.2   &  74.0     &  75.4     &  70.9     & 85.1      & 76.8 \\
    Ours (Latin) &   77.8    &  74.9     &  75.8     &  71.3     &      84.9 &  77.5 \\
    GCAN (FT) & 78.8  & 76.4  & 78.4  & 74.1  & 86.5  & 79.2  \\
    Ours (FT) &  80.0   &  78.0     &  78.5     &  74.9     & 86.6      & 80.1 \\
    \bottomrule
    \end{tabular}%
 
      \vspace{-20pt}
\end{table}

\begin{table}[htpb]
    \caption{Recognition accuracy on the previous Latin benchmark. The performance of GCAN is reported in \cite{qiao2021gaussian}.}
    \label{table:Latin_benchmark}
  \centering
    \begin{tabular}{c|ccccccc}
    \toprule
          & IIIT5K & SVT   & IC13  & IC15  & SVTP  & CT80  & All \\
    \midrule
    GCAN  & 94.4  & 90.1  & 93.3  & 77.1  & 81.2  & 85.6  & 87.8  \\
    Ours  & 95.3  & 90.1  & 93.0  & 79.1  & 82.6  & 89.6  & 88.9  \\
    \bottomrule
    \end{tabular}%

    \vspace{-20pt}
\end{table}%

\section{Compatibility with Text Detectors}

\begin{figure}[htpb]
    \centering
    \includegraphics[width=\textwidth]{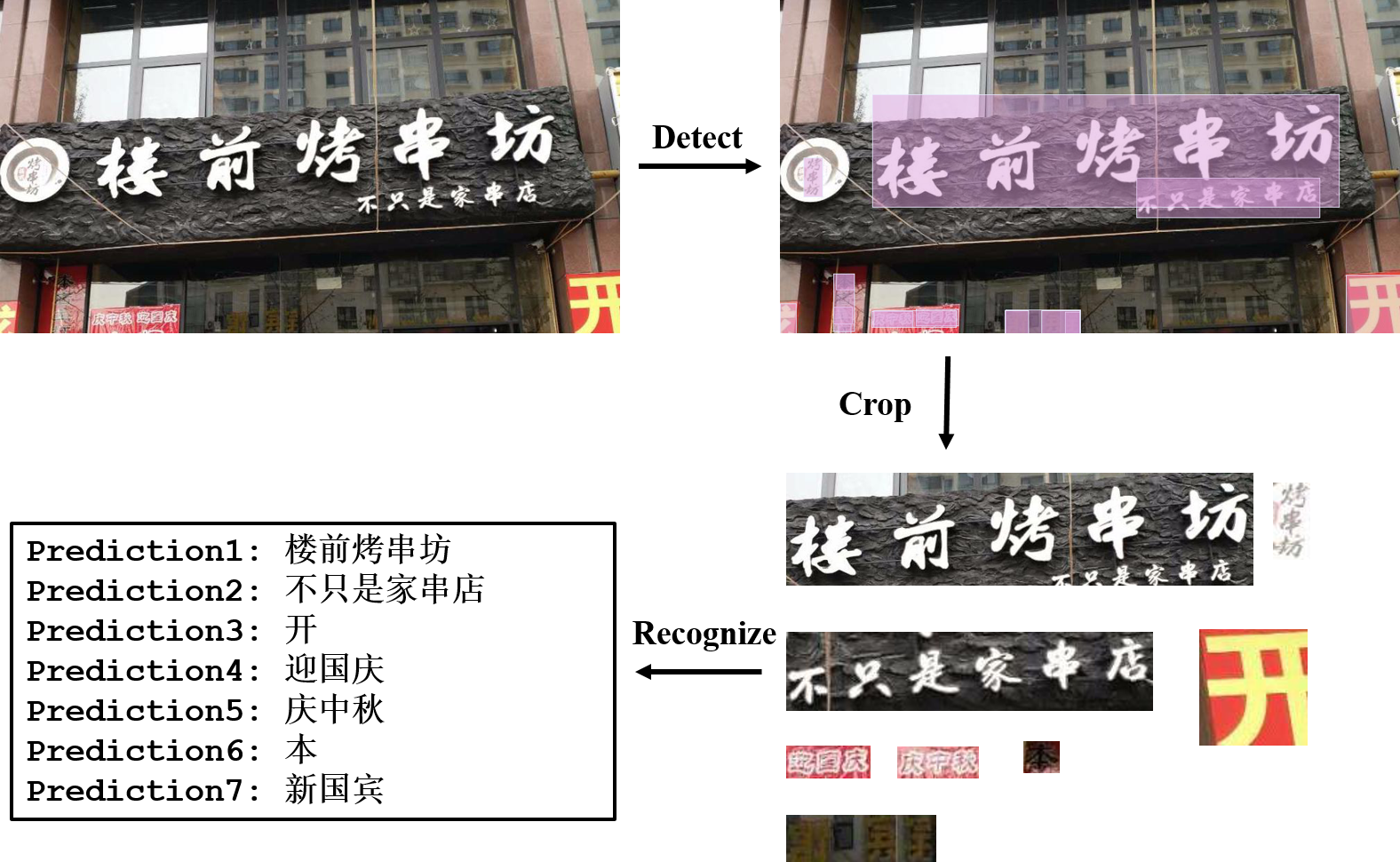}
    \caption{Our text spotting system.}
    \label{fig:spotting_system}
    \vspace{-15pt}
\end{figure}

The scene text spotting is to localize and recognize text boxes from panoramic images, and the recognizer is a downstream processor of the detector. The upstream detector may generate bounding boxes of all sizes, and the border of these boxes may not cover ground truth boxes exactly. In order to evaluate our model in the realistic setting, we employ the box discretization network \cite{liu2019omnidirectional, liu2019exploring} as our detector and use our recognition model as the downstream recognizer. We give an example of the text spotting results on ReCTS test datasets in Fig. \ref{fig:spotting_system}, and the end-to-end text spotting performance is shown in Tab. \ref{tab:spotting_performance}. The performance of the text detector is evaluated by \textit{recall, precision} and \textit{hmean}, and \textbf{1-NED} is to evaluate the performance of the recognition performance. It is worth noting that we do not utilize the ensemble tricks during the detection stage, so the detection performance is slightly below the results given by \cite{liu2019omnidirectional, liu2019exploring}. The results shown in Tab. \ref{tab:spotting_performance} validate the compatibility of our model with the text detector, and demonstrate superiority to the baseline (GCAN).

\begin{table}[htpb]
    \caption{End-to-end performance on the ReCTS-19 test dataset.}
    \label{tab:spotting_performance}
    \centering
    \begin{tabular}{c|ccc|c}
    \toprule
    Models & Recall & Precision & Hmean & 1-NED \\
    \midrule
    GCAN  & \multirow{2}{*}{87.9 } & \multirow{2}{*}{95.7 } & \multirow{2}{*}{91.6 } & 78.6  \\
    Ours  &       &       &       & 79.0  \\
    \bottomrule
    \end{tabular}%

    \vspace{-5pt}
\end{table}

\section{A Demonstration of BCE Loss Masks}
\begin{figure}
    \centering
    \includegraphics[width=\textwidth]{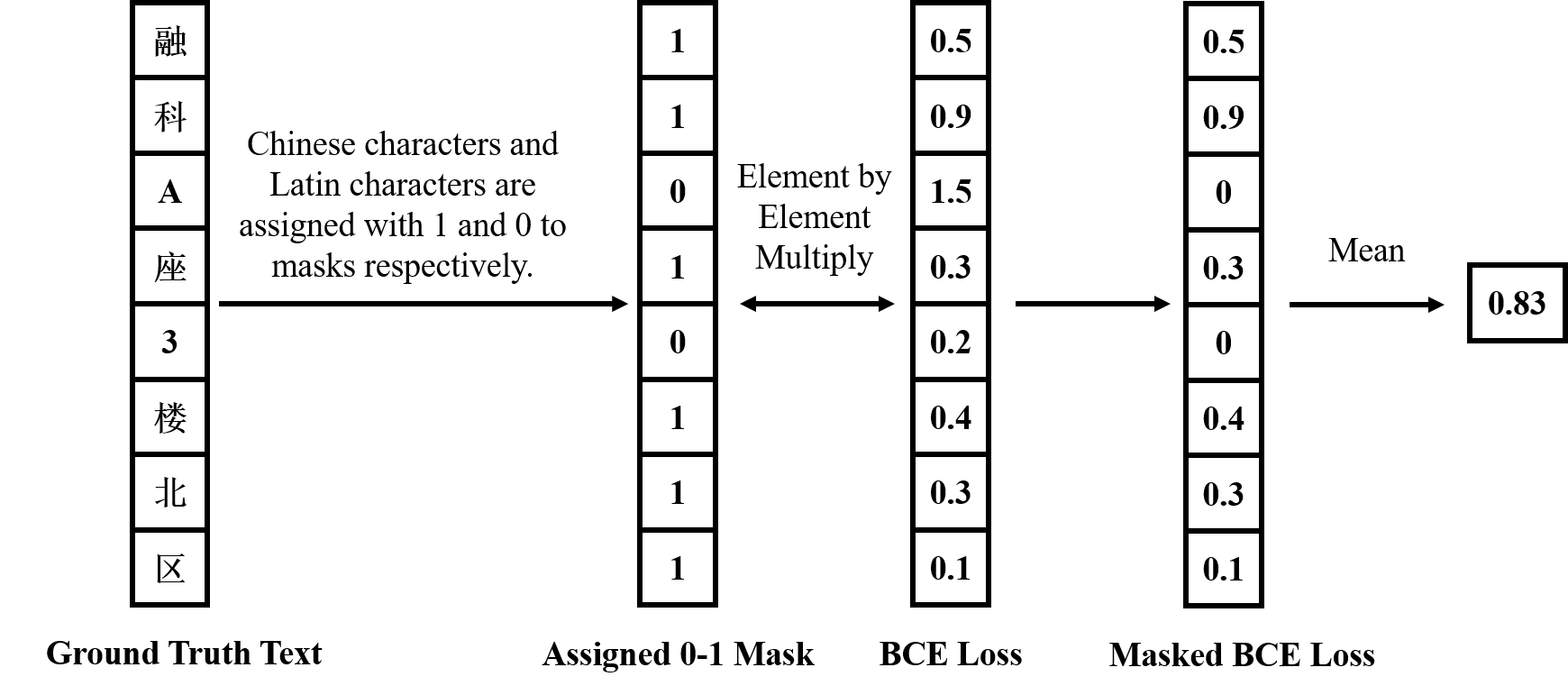}
    \caption{The figure shows how we assign 0-1 mask for the BCE loss of each word. The Latin characters are not taken into the computation of BCE loss.}
    \label{fig:BCE_Mask}
\vspace{-5pt}
\end{figure}

Since there are no radicals in Latin characters, we use a mask to label zero for the them and one for Chinese characters, and the mask is multiplied by the loss matrix, as shown in Fig.\ref{fig:BCE_Mask}.

\section{Discussions about Related Chinese OCR Work}
In this section, we give a discussion about the differences between our use of Chinese radicals and previous papers.
\begin{figure}[htpb]
    \centering
    \includegraphics[width=\textwidth]{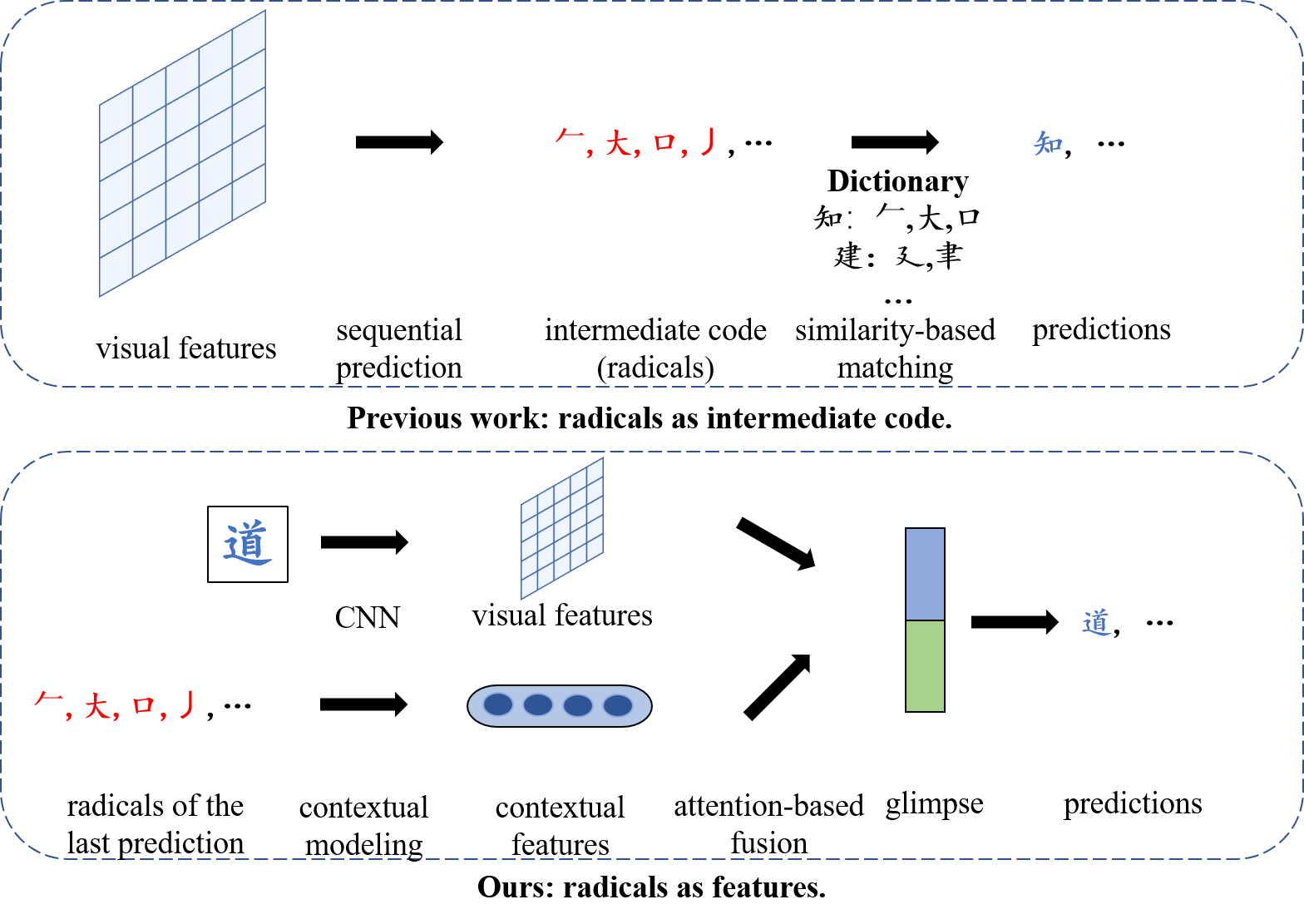}
    \caption{The top figure demonstrates the previous work that takes radicals as a kind of intermediate code for character matching, and the bottom shows how we use radicals as a kind of features for classification.}
    \label{fig:my_label}
\end{figure}

In most of the related work \cite{zhang2018radical, wang2018denseran, wu2019joint, jaderberg2015spatial, wang2019radical, wang2017radical}, Chinese radicals are taken as a kind of intermediate code (IC), while we use it as a kind of embedding to fuse with visual features. Hence there are fundamental differences between them. We use Fig. \ref{fig:my_label} to compare these two different methodology.

\begin{figure}
    \centering
    \includegraphics[width=\textwidth]{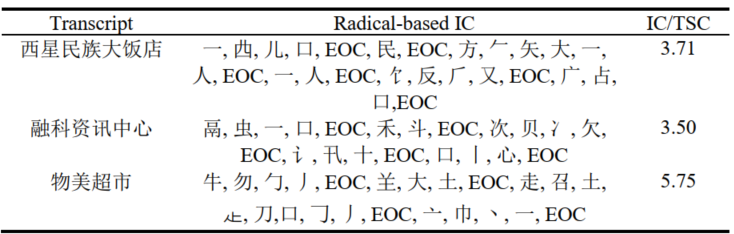}
    \caption{Some examples of radical-based intermediate code (IC). We calculate the ratio of the length of IC to the length of transcripts (IC/TSC), and find it is approximately 4 $\sim$ 5. EOC means \textit{end of a character}.}
    \label{fig:IC_examples}
\end{figure}

For {single-character recognition}, using radicals as IC is feasible. However, for a sequence of characters, the code would be very long, making it impractical for recognition. As shown in Fig. \ref{fig:IC_examples}, a transcript usually generates 4x $\sim$ 5x the length of radical-based IC. In STR datasets, transcripts with 10 $\sim$ 20 characters are common, which means the length of IC can reach 50 $\sim$ 100.

Hence, we conclude that long radical-based or stroke-based intermediate code may (1) make sequential modeling more difficult; (2) increase the time complexity. In contrast to radical-based IC, our method is not influenced by the length of transcripts since they are translated into radical-embedding with the same dimension and fused with visual features. As a result, our model benefits from low time complexity and shorter sequence to prediction.

\newpage

\bibliography{egbib}